\documentclass[letterpaper, 10 pt, journal, twoside]{./iEEEtran}
\usepackage{amsmath,amssymb,amsfonts}
\usepackage{balance}
\usepackage{bm}
\usepackage{hyperref}
\usepackage{color}
\usepackage[table,xcdraw]{xcolor}
\usepackage{tabularx}
\usepackage[caption=false,font=normalsize,labelfont=sf,textfont=sf]{subfig}
\usepackage{booktabs}
\usepackage{graphicx}
\usepackage{multirow}
\usepackage{tikz}
\usepackage{pgfplots}
\usepackage{siunitx}
\usetikzlibrary{bending, external}

\graphicspath{{./Figures/}}
\DeclareGraphicsExtensions{.pdf,.png,.jpg,.eps,.svg}

\IEEEoverridecommandlockouts
\pgfplotsset{compat=1.16} 
\title{COMET: A Dual Swashplate Autonomous Coaxial Bi-copter AAV with High-Maneuverability and Long-Endurance}
\author{
    Shuai Wang\textsuperscript{1}, 
    Xiaoming Tang\textsuperscript{1}, 
    Junning Liang\textsuperscript{1}, 
    Haowen Zheng\textsuperscript{1},
    Biyu Ye\textsuperscript{1},
    Zhaofeng Liu\textsuperscript{1},
    Fei Gao\textsuperscript{3,4},
    Ximin Lyu\textsuperscript{1,2,4}

	\thanks{Manuscript received: Jul, 22th, 2025; Revised Sep, 17th, 2025; Accepted Nov, 12th, 2025. This paper was recommended for publication by Editor Giuseppe Loianno upon evaluation of the Associate Editor and Reviewers' comments.
    (\textit{\textbf{Shuai Wang, Xiaoming Tang, and Junning Liang contributed equally to this work.}})
    }
    \thanks{This work was supported by the National Natural Science Foundation of China (Grant No. 62303495), the Young Talent Support Project of Guangzhou Association for Science and Technology (Grant No. QT-2025-004), and the National Key Research and Development Program of China (Grant No.2023YFB4706600). (Corresponding author: Ximin Lyu.)}
      \thanks{$^1$the School of Intelligent Systems Engineering, Sun Yat-sen University, Guangzhou 510275, China (e-mail: \{\textit{tangxm7, liangjn33, zhenghw23, yeby9, liuzhf36}\}\textit{@mail2.sysu.edu.cn}, \{\textit{wangsh368, lvxm6}\}\textit{@mail.sysu.edu.cn}).
      }
      \thanks{$^2$the Research Institute of Multiple Agents and Embodied Intelligence, Peng Cheng Laboratory, Shenzhen 518055, China.}
      \thanks{$^3$State Key Laboratory of Industrial Control Technology, Zhejiang University, Hangzhou, China. (e-mail: \textit{fgaoaa@zju.edu.cn})}
      \thanks{$^4$Differential Robotics Technology Co., Ltd., Hangzhou, China.}
     \thanks{Digital Object Identifier (DOI): see top of this page.}
}
\begin{document}
\maketitle
\begin{abstract} 
Coaxial bi-copter autonomous aerial vehicles (AAVs) have garnered attention due to their potential for improved rotor system efficiency and compact form factor.
However, balancing efficiency, maneuverability, and compactness in coaxial bi-copter systems remains a key design challenge, limiting their practical deployment.
This letter introduces COMET, a coaxial bi-copter AAV platform featuring a dual swashplate mechanism.
The coaxial bi-copter system's efficiency and compactness are optimized through bench tests, and the whole prototype's efficiency and robustness under varying payload conditions are verified through flight endurance experiments.
The maneuverability performance of the system is evaluated in comprehensive trajectory tracking tests.
The results indicate that the dual swashplate configuration enhances tracking performance and improves flight efficiency compared to the single swashplate alternative.
Successful autonomous flight trials across various scenarios verify COMET's potential for real-world applications.
\end{abstract}


\section{Introduction}
\IEEEPARstart{M}{icro} aerial vehicles (MAVs), particularly small-scale rotary-wing AAVs, have gained widespread adoption in applications such as environmental monitoring~\cite{environmental_monitoring}, disaster response~\cite{disaster}, indoor exploration~\cite{exploration}, and infrastructure inspection~\cite{infrastructur_inspection}. 
Their vertical takeoff and landing (VTOL) capability, compact structure, and agile maneuverability make them well-suited for autonomous operations in GPS-denied or confined environments~\cite{uav_2018}.
Despite these advantages, small rotary-wing AAVs face fundamental challenges in energy efficiency and endurance. 
Constrained by limited horizontal projection area, they typically rely on multiple small-diameter rotors, which reduce the thrust-to-power ratio and result in high energy consumption per unit thrust~\cite{Koe2012single, Koehl_Rafaralahy_Martinez_Boutayeb_2010}. 
These limitations critically constrain flight time, particularly in high-payload or long-duration missions.

Enhancing rotor system efficiency while maintaining a compact horizontal projection area is therefore crucial for advancing MAV endurance performance. 
Under such constraints, coaxial bi-copter systems offer a promising solution~\cite{buzzatto2022benchmarking}. 
By stacking two counter-rotating rotors along a shared vertical axis, coaxial bi-copter designs effectively increase the rotor disk area without enlarging the lateral dimension. 
This architecture enables high power density and compactness, making it particularly suitable for small rotary-wing AAVs targeting extended endurance and portability.

The rotor system of a coaxial bi-copter AAV is tasked with providing the main thrust for the vehicle while simultaneously achieving full attitude control. This dual responsibility presents a significant challenge to the design and implementation of coaxial bi-copter systems.
Thus, it is hard to balance efficiency, maneuverability, and compactness in coaxial bi-copter systems without considering all factors systematically.

\begin{figure*}[t]
\centering
\includegraphics[width=0.95\linewidth]{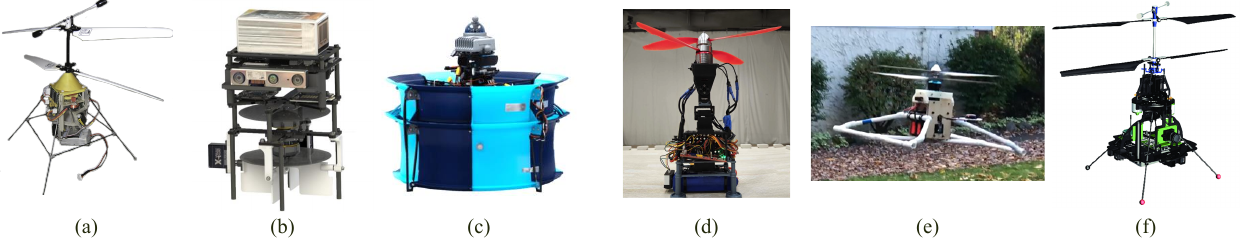}
\caption{(a).~A center-of-gravity (CoG) shifting driven coaxial bi-copter AAV muFly by Bermes~\textit{et al}.~\cite{bermes2008CoG}
(b).~Aerodynamic control surfaces coaxial bi-copter AAV CanFly by Pan~\textit{et al}.~\cite{canfly}
(c).~A swashplateless duct coaxial bi-copter AAV Halo by Li~\textit{et al}.~\cite{li2023halo}
(d).~A servo-controlled coaxial bi-copter AAV by Chen~\textit{et al}.~\cite{chen2024design}
(e).~A gimbal-controlled coaxial bi-copter AAV GimbalHawk by Reddington~\textit{et al}~\cite{reddington2021design}. 
(f).~A coaxial bi-copter AAV CoAX with swashplate cyclic pitch control by Fankhauser~\textit{et al}.~\cite{fank2011}}
\label{fig:otheruav}
\vspace{-10pt}
\end{figure*}

In this work, we present \textbf{COMET} (Coaxial Bi-copter AAV with High-Maneuverability and Long-Endurance).
COMET integrates a coaxial bi-copter system with a dual swashplate cyclic pitch mechanism.
The dual swashplate cyclic pitch mechanism enables the independent actuation of both rotors, providing improved agility.
The rotor system is optimized through bench tests, which define the optimal rotor separation distance and installation angle, ensuring optimal rotor system efficiency while maintaining compactness.
Comparative experiments in endurance and maneuverability validate the effectiveness of our system.

The main contributions of this work are as follows:
\begin{itemize}
\item \textbf{Design and implementation of COMET:} We introduce a coaxial bi-copter AAV equipped with a dual swashplate cyclic pitch control system, enabling independent actuation of the upper and lower rotors to significantly enhance maneuverability.

\item \textbf{Optimization of the coaxial bi-copter system:} We conduct systematic testbench experiments to jointly optimize rotor separation distance and blade installation angle, improving rotor system efficiency while preserving a compact vertical profile.

\item \textbf{Comprehensive experimental validation:} We validate the endurance of COMET’s rotor system under varying load conditions, the necessity of the dual swashplate configuration for dynamic flight, compared to the single swashplate setup, and the autonomous capabilities in unknown environments through extensive and sufficient experiments.

\end{itemize}

\begin{figure*}[t]
    \centering
    \includegraphics[width=\linewidth]{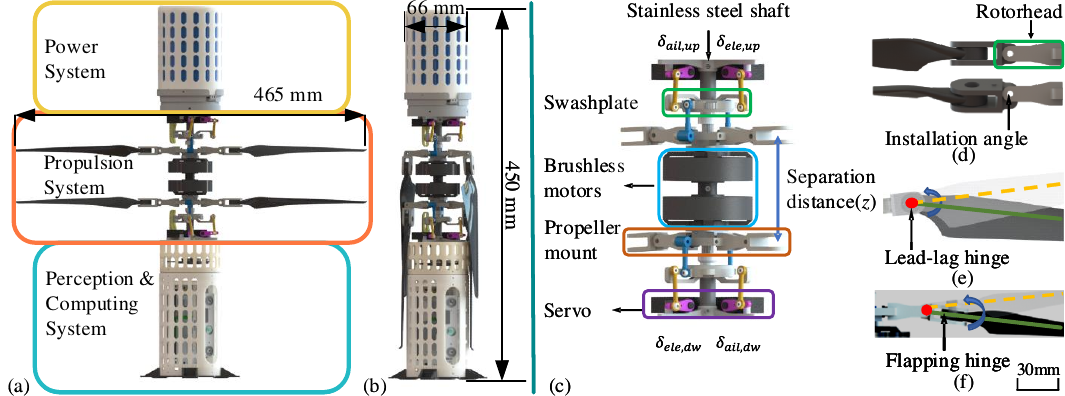}
    \caption{(a).~General schematic of COMET. (b).~The folded overall dimensions of COMET are 450~mm in height and a maximum diameter of 66~mm. (c).~Breakdown of the COMET's propulsion system, highlighting key structural components including the swashplates, brushless motors, propeller mount, and servos. The diagram also illustrates the vertical rotor separation distance $z$, (d).~Blade installation angle, (e).~Flapping hinges, and (f).~Lead-lag hinges.}
    \label{fig:overall}
\end{figure*}

\section{Related Work}
\subsection{Actuation Strategies}
The actuation mechanism is a critical factor in the design and performance of coaxial bi-copter systems. 
To achieve effective and agile attitude control, researchers have explored various approaches, including center-of-gravity (CoG) shifting~\cite{boua2006CoG, bermes2008CoG, wang2006CoG}, aerodynamic control surfaces~\cite{canfly, deng2020aerodynamic, dominguez2022micro}, swashplateless actuation~\cite{li2023halo, Paulos2018}, servo or gimbal-based thrust vectoring~\cite{reddington2021design, chen2024design}, and cyclic pitch control via swashplates~\cite{Koe2012single, fank2011, gun_2013}.

MuFly~\cite{bermes2008CoG}, shown in Fig.~\ref{fig:otheruav}(a), employs a CoG-shifting mechanism that repositions onboard masses (e.g., batteries) to induce attitude changes. 
This approach enables basic attitude control in micro AAVs with inherently passive stability. 
However, due to limited movable mass and short moment arms, CoG shifting provides poor control torque and poor agility, particularly in larger (above 250~g take-off weight) or long-endurance platforms.
Canfly~\cite{canfly} uses aerodynamic control surfaces (e.g., rudders, fins) to deflect rotor downwash airflow to generate aerodynamic forces, as shown in Fig.~\ref{fig:otheruav}(b). 
However, due to the limited size of the rotor, the aerodynamic control force generated by the deflection of the control surfaces in the downstream flow is relatively limited and can be easily affected by crosswinds.
HALO~\cite{li2023halo}, shown in Fig.~\ref{fig:otheruav}(c), features a coaxial ducted structure and employs a swashplateless design to reduce mechanical complexity. 
It achieves three-axis attitude control using only two motors by leveraging differential speed modulation and passive flapping dynamics of the blades.
However, the reliance on frequent motor speed changes introduces current ripple, which can increase thermal load and potentially degrade both efficiency and long-term reliability~\cite{qin_2024}.
Chen \textit{et al}.~\cite{chen2024design} propose a servo-actuated coaxial bi-copter AAV, whereas GimbalHawk~\cite{reddington2021design} employs a gimbal-based mechanism, as illustrated in Fig.~\ref{fig:otheruav}(d) and (e), respectively.
However, tilting the entire rotor assembly introduces additional gyroscopic moments, which complicate control. 
Moreover, the minimal rotor separation exacerbates aerodynamic interference, limiting overall system efficiency.
Swashplate-based cyclic pitch control systems are widely used in micro autonomous helicopters, including coaxial bi-copter AAVs. 
These systems can be configured with either a single swashplate or a dual swashplate.
CoAX~\cite{fank2011} shown in Fig.~\ref{fig:otheruav}(f) adopts a single swashplate configuration derived from commercial model helicopter designs.
However, unlike single swashplate systems, dual swashplate architectures allow independent control of each rotor’s cyclic pitch, significantly enhancing control authority and increasing response bandwidth for agile attitude maneuvers~\cite{zhao_2022}.

Previous studies~\cite{KimBrown2008} show that compact aerial vehicles with coaxial rotors inherently benefit from induced power advantages over single-rotor systems, especially in hover and at low advance ratios. Swashplate-based cyclic pitch control, through thrust-vector tilting and flapping dynamics, is demonstrated to enhance maneuverability in coaxial bi-copter AAVs~\cite{fank2011}. 
Additionally, swashplate-based cyclic pitch control is proven to improve efficiency and control response in small-scale AAVs~\cite{SwashplateControlUAV}, further supporting its use in coaxial bi-copter systems.

\subsection{Rotor System Optimization}
The primary challenge in coaxial bi-copter system design stems from airflow interference between the rotors. 
This interference introduces substantial aerodynamic disturbances, negatively impacting overall system performance.
Mitigating such effects hinges on optimizing two critical rotor design parameters: (1).~the vertical separation distance between rotors, and (2).~the blade installation angle.

The vertical spacing between the upper and lower rotors is a primary factor influencing wake interaction, as identified in prior works~\cite{deng2020aerodynamic,fank2011,Rama2013aero,Hai2018aero,lak2010areo,ramasamy2015hover}.
Ramasamy \textit{et al.}~\cite{Rama2013aero,ramasamy2015hover} experimentally demonstrate that inadequate vertical spacing leads to severe wake impingement on the lower rotor, reducing its effective lift and increasing the power required to maintain thrust equilibrium. 
Buzzatto \textit{et al.}~\cite{buzzatto2022benchmarking} develop an open-source testbench to benchmark coaxial rotor systems and systematically analyze the relationship between mechanical and electrical power with respect to rotor separation ratio across varying rotor diameters.
The blade installation angle, including pitch and coning angles, also plays a critical role in influencing rotor–rotor aerodynamic interaction in coaxial systems. 
Hai \textit{et al.}~\cite{Hai2018aero} note that slight variations in installation angles can significantly alter the pressure distribution across rotor planes, affecting stability and efficiency. 
 Additionally, Lakshminarayan and Baeder~\cite{lak2010areo} show through computational fluid dynamics that carefully adjusting blade pitch profiles promotes smoother inflow and reduces local flow separation in coaxial configurations, thereby improving aerodynamic performance.

Together, these findings highlight the importance of jointly optimizing rotor separation and blade installation geometry to minimize interference effects, thereby ensuring optimal rotor system efficiency while maintaining compactness.

\section{System Design, Optimization and Implementation}
\subsection{System Overview and Propulsion Architecture}
\label{sec:overall_design}
COMET integrates three core subsystems: the power system, the propulsion system, and the perception and computing system, as illustrated in Fig.~\ref{fig:overall}(a).
The whole system height of the COMET is 450~mm and approximately 66~mm in diameter in the folded state, as shown in Fig.~\ref{fig:overall}(b).
The power system supports interchangeable battery packs with varying capacities to accommodate diverse mission requirements, including 3000 mAh and 5000 mAh 6s Li-ion battery options.
The perception and computing system comprises an NVIDIA Orin NX onboard computer for high-level planning and decision-making, and a Holybro Kakute H7 flight control unit for low-level control execution.
Onboard sensing is provided by either an Intel RealSense D430 stereo camera or a Livox Mid-360 LiDAR, depending on the mission requirements.
The propulsion system is based on a coaxial bi-copter configuration, with two DT-5008 brushless DC (BLDC) motors, two-pair propeller mounts with 16x6 two-blade carbon composite propellers, and two swashplates.
All assemblies are mounted on a stainless steel shaft, as shown in Fig.~\ref{fig:overall}(c).
The elevator and aileron servo inputs for the upper and lower swashplates are denoted as $\delta_{\mathrm{ele}, up}$ and $\delta_{\mathrm{ail}, up}$, $\delta_{\mathrm{ele}, dw}$ and $\delta_{\mathrm{ail}, dw}$, with identical control signals, i.e., $\delta_{\mathrm{ele}, up} = \delta_{\mathrm{ele}, dw}$, and $\delta_{\mathrm{ail}, up} = \delta_{\mathrm{ail}, dw}$.
The rotorhead combines flapping and lead–lag hinges, producing a compact assembly with favorable rotor dynamics. 
The blades are mounted onto the propeller mounts with a specific installation angle via lead-lag hinges, as shown in Fig.~\ref{fig:overall}(d).
The lead–lag hinges, shown in Fig.~\ref{fig:overall}(e), permit in-plane blade motion, relieving the asymmetric aerodynamic and inertial loads during aggressive maneuvers, enhancing flight stability~\cite{wang_2012_dynamic}.
The propeller mounts are connected to the rotorhead through flapping hinges, enabling out-of-plane flapping motion around the hinge axis, as illustrated in Fig.~\ref{fig:overall}(f).
COMET has a projected diameter of 465 mm, with blades folding downward for compact storage, reducing the platform's maximum diameter to 66 mm and achieving a 98.0\% reduction in projected area.
The overall components of COMET are shown in Table.~\ref{tab:components}.

To verify COMET's structural robustness, we conducted finite element modal simulations to analyze structural modes and performed Fast Fourier Transform (FFT) using IMU data during hover flight to identify vibration modes.
The results show that the actual vibration mode at 23 Hz is close to the simulated modes (28.3 Hz and 31.5 Hz), all well separated from the 45 Hz rotor frequency, ensuring no structural resonance. 
Cumulative flight tests over 100 hours confirm no performance degradation, excessive vibration, or structural failures, supporting the robustness and reliability of COMET's propulsion system and design.

\begin{table}[h]
  \centering
  \caption{Weight and Model of Each Component}
  \label{tab:components}
  \renewcommand{\arraystretch}{1.5}
  \begin{tabular}{p{1.2cm} p{2.2cm} p{2.6cm} >{\centering\arraybackslash}p{1.2cm}}
    \specialrule{0.2em}{0em}{0.2em} 
    \textbf{Category} & \textbf{Component} & \textbf{Model} & \textbf{Weight (g)} \\
    \midrule
    \multirow{2}{2.6cm}{Power\\ System} 
    & Battery 1 & 5000mAh 21700 & 480 \\
    & Battery 2 & 3000mAh 18650 & 340 \\
    \midrule
    \multirow{3}{2.6cm}{Propulsion\\ System} 
    & Propeller & EOLO CN16x6 & 58 \\
    & Motor & DT-5008 & 245 \\
    & Servo & BlueArrow BA8-3 & 6.8 \\
    \midrule
    \multirow{5}{2.6cm}{Perception\\ Computing\\ System} 
    & Flight Controller & Holybro Kakute H7 & 8 \\
    & ESC & Holybro Tekko32 50A & 8 \\
    & Onboard Computer & NVIDIA Orin NX & 80 \\
    & Stereo Camera & Intel RealSense D430 & 30 \\
    & LiDAR (Optional) & Livox Mid-360 & 300 \\
    \specialrule{0.2em}{0.2em}{0em}
  \end{tabular}
\end{table}

\subsection{Propulsion System Design And Optimization}
The propulsion system is a central element of COMET’s mechanical design, engineered to achieve compact integration within a limited vertical envelope while enhancing efficiency under the premise of a compact longitudinal structure.
The vertical separation distance $z$ between the upper rotor and lower rotor, as well as the installation angle of each blade, are critical parameters that directly affect rotor–rotor interaction, lift symmetry, and overall control efficiency. 
An optimized vertical separation distance $z$ can mitigate aerodynamic interference between the rotors while influencing the system’s center of gravity and thrust allocation. 
Similarly, a carefully adjusted installation angle can improve lift distribution and enhance control responsiveness. 

\begin{figure}[h]
    \centering
    \includegraphics[width=\linewidth]{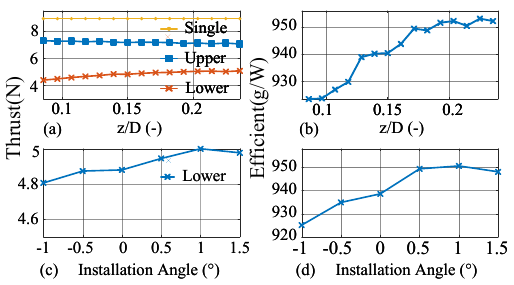}
    \caption{(a)(b) Thrust and efficiency versus normalized separation distance ($z/D$). 
(c)(d) Thrust and efficiency versus installation angle of the lower rotor.}
    \label{fig:dis_result}
\end{figure}

We experimentally evaluate these parameters to quantify their effects on propulsion efficiency and guide the final design configuration of COMET.
The thrust generated by the motor was measured using an SRI M3813B high-precision force and torque sensor. 
The sensor offers a maximum force error of 0.04 N and a maximum torque error of 0.009 N·m within its rated measurement range.
The motors are rigidly mounted on the force sensor, and the sensors themselves are securely fixed to the test rig.
During the experiment, both motors are operated at equal torque, with the total throttle fixed at 40\% to replicate COMET's hover condition.
As shown in Fig.~\ref{fig:dis_result}(a), with the increase in the normalized separation distance $z/D$, where $D$ is the rotor diameter, the thrust generated by the lower rotor also increases. 
The result suggests that increasing the vertical spacing effectively mitigates rotor–rotor interaction.
Consequently, the overall efficiency of the system gradually improves and tends to stabilize when the separation ratio reaches $z/D = 0.17$, which corresponds to a separation of 79~mm, as illustrated in Fig.~\ref{fig:dis_result}(b).
Since further increasing $z/D$ yields no significant improvement in system efficiency, $z=79~mm$ is selected as the rotor system design parameter to maintain structural compactness.
Installation angle tests are conducted at the fixed optimal separation distance of 79~mm. 
As the lower rotor is more exposed to the aerodynamic wake of the upper rotor, optimizing its installation angle is likely to produce a greater improvement in thrust efficiency~\cite{Giljarhus_2022}. 
Therefore, only the lower rotor’s installation angle is tested, ranging from~–1° to +1.5°, at 0.5° intervals.
As shown in Fig.~\ref{fig:dis_result}(c) and (d), both the lower rotor thrust and the overall system efficiency increase with the installation angle.
Thrust magnitude and efficiency are both maximized at $+1^\circ$, and then begin to decline.
Therefore, we chose $+1^\circ$ as the installation angle for the lower blades.

\begin{figure}[h]
    \centering
    \includegraphics[width=0.75\columnwidth]{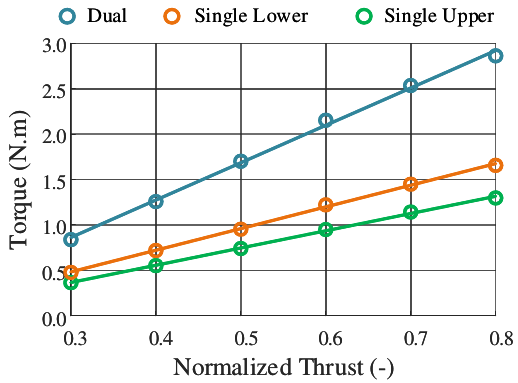}
    \caption{Testbench torque results, blue for dual swashplate configuration, orange for single swashplate configuration on lower rotor, green for single swashplate configuration on upper rotor.}
    \label{fig:ft_result}
\end{figure}

To verify the effectiveness of the COMET dual swashplate configuration in providing control torque compared to the single swashplate configuration, we conduct testbench experiments to measure the differences in maximum control torque between the single swashplate and dual swashplate configurations. 
We recorded the average of the maximum and minimum values of the control torque generated by cyclic pitch and its trend as the normalized thrust changed, as shown in Fig.~\ref{fig:ft_result}.
The results indicate that the maximum control torque provided by the COMET cyclic pitch mechanism is related to the normalized thrust and varies linearly with it. 
The maximum control torque of the COMET with the dual swashplate configuration is higher than that of both single swashplate configurations.
Compared to the single swashplate configuration on the upper rotor, the dual swashplate configuration increases the control torque by 116.7\%, and compared to the single swashplate configuration on the lower rotor, the dual swashplate configuration increases the control torque by 72.51\%.
The results demonstrate the effectiveness of the dual swashplate configuration in enhancing the maximum control torque.

\section{Modeling and Controller Design}
\begin{figure}[h]
    \centering
    \includegraphics[width=0.85\linewidth]{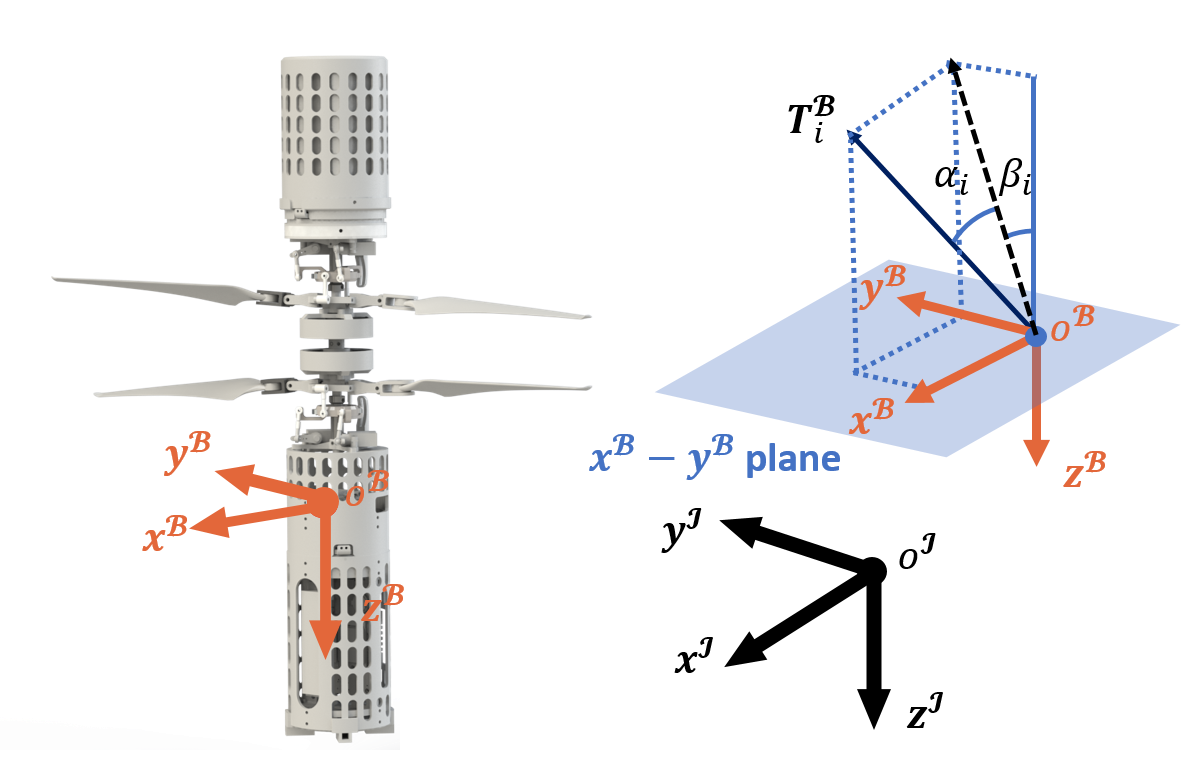}
    \caption{Reference frame definition of COMET and visualization of the tilted thrust vector with tilt angles $\alpha_i$ and $\beta_i$. }
    \label{fig:frame_config}
    \vspace{-10pt}
\end{figure}

\subsection{Dynamics}
We introduce two coordinate frames of the proposed system for further discussion and they are shown in Fig.~\ref{fig:frame_config}: the body frame $\bm{x}^{\mathcal{B}}$, $\bm{y}^{\mathcal{B}}$, $\bm{z}^{\mathcal{B}}$, and the inertial frame $\bm{x}^{\mathcal{I}}$, $\bm{y}^{\mathcal{I}}$, $\bm{z}^{\mathcal{I}}$. 
The body frame's origin $o^{\mathcal{B}}$ is set to coincide with the vehicle's center of gravity (CoG). 
We formulate the equation of motion for the COMET as follows
\begin{equation}
    \begin{aligned}
        m\dot{\bm{v}}^{\mathcal{I}} &= m\bm{g} + \mathbf{R}(\bm{F}^{\mathcal{B}} + \bm{F}^{\mathcal{B}}_{ext}), \\
        \mathbf{J}\dot{\bm{\omega}}^{\mathcal{B}} &= \bm{M}^{\mathcal{B}} - \bm{\omega}^{\mathcal{B}}{\times}\mathbf{J}\bm{\omega}^{\mathcal{B}} + \bm{\tau}^{\mathcal{B}}_{ext},
    \end{aligned}
\end{equation}

where $m$ denotes the AAV's total mass, $\bm{v}^{\mathcal{I}}$ is the vehicle's velocity vector in the inertial frame, $\bm{g}$ represents the gravitational vector, $\mathbf{R}$ is the rotation matrix from the body frame to the inertial frame, $\bm{F}^{\mathcal{B}}$ is the force generated by the rotors in the body frame, $\bm{F}^{\mathcal{B}}_{ext}$ is the collection of the rotor drag and fuselage drag in the body frame, $\mathbf{J}$ is the inertia matrix, and $\bm{\omega}^{\mathcal{B}}$ and $\bm{M}^{\mathcal{B}}$ are the angular rate and net torque in the body frame, respectively. 
Additionally, $\bm{\tau}^{\mathcal{B}}_{ext}$ represents the external torque acting on the vehicle's CoG, induced by rotor and fuselage drag.
Rotor thrusts, denoted by $\bm{T}^{\mathcal{B}}_i$ for $i \in \{\textnormal{up}, \textnormal{dw}\}$, where ``$\textnormal{up}$'' refers to the upper rotor and ``$\textnormal{dw}$'' refers to the lower rotor, are oriented perpendicular to their respective Tip Path Planes (TPPs).
The deviation of the TPP can be described by the longitudinal flapping angle ${\alpha}_i$ and the lateral flapping angle ${\beta}_i$, so the rotor thrust vectors are
\begin{equation}
\bm{F}^{\mathcal{B}} = \sum{\vert\bm{T}^{\mathcal{B}}_i\vert}
\begin{bmatrix}
-\sin ({\alpha}_i) \\
\sin ({\beta}_i) \\
-\cos ({\alpha}_i) \cos ({\beta}_i)
\end{bmatrix}.
\end{equation}
  
When the TPPs deviate from their nominal orientation, the thrust vectors no longer intersect the center of gravity (CoG). 
This misalignment generates off-axis moments about the CoG, expressed as  $\bm{l}_{\textnormal{up}} \times \bm{T}^{\mathcal{B}}_{\textnormal{up}}$ and $\bm{l}_{\textnormal{dw}} \times \bm{T}^{\mathcal{B}}_{\textnormal{dw}}$, where $\bm{l}_{\textnormal{up}}$ and $\bm{l}_{\textnormal{dw}}$ are the displacement vectors from the CoG to the upper and lower rotor hubs. 
The deviation of the TPP also directly results in flapping torques. 
The resulting flapping-induced torques are modeled as a linear torsional spring proportional to the rotor's thrust, with a stiffness constant $K_{\beta}$, based on the testbench results from Fankhauser~\textit{et al}.~\cite{fank2011} under fixed thrust and our testbench results shown in Fig.~\ref{fig:ft_result}.
The combined torque is expressed as
\begin{equation} 
\label{eq:combind_moment}
\bm{M}^{\mathcal{B}}
= \sum \bm{l}_i \times \bm{T}^{\mathcal{B}}_i
+ \sum K_\beta
\begin{bmatrix}
\sin ({\alpha}_i) \\
\sin ({\beta}_i) \\
0
\end{bmatrix}
\vert\bm{T}^{\mathcal{B}}_i\vert.
\end{equation}
In COMET, the rotor swashplate mechanism is configured with a flapping phase lag of $45^\circ$, effectively decoupling longitudinal and lateral flapping dynamics~\cite{wang_2012_dynamic}, simplifying the mapping between servo commands and TPP orientation to
\begin{equation}
{\alpha}_i = A_i \delta_{\textnormal{ele}}, \qquad {\beta}_i = B_i \delta_{\textnormal{ail}},
\end{equation}
where $\delta_{\mathrm{ele}}$ and $\delta_{\mathrm{ail}}$ are the normalized upper and lower elevator and aileron command inputs, ranging from $[-1, 1]$, $A_i$ and $B_i$ represent the ratios between the normalized control inputs and the TPP flapping angles, respectively.

\subsection{Control System}
\begin{figure}[h]
    \centering
    \includegraphics[width=\columnwidth]{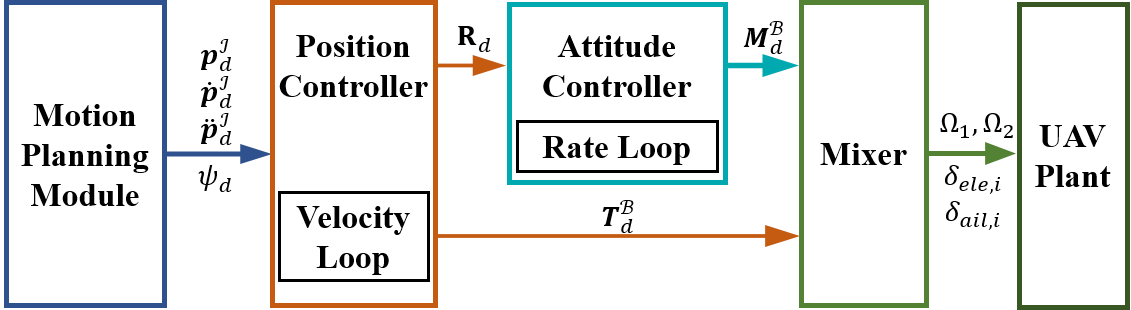}
    \caption{Control Structure for COMET}
    \label{fig:control_structure}
\end{figure}

The whole control architecture of the proposed system is illustrated in Fig.~\ref{fig:control_structure}. 
The motion planning module generates target positions and their first and second derivatives $\bm{p}_d^{\mathcal{I}}, \dot{\bm{p}}_d^{\mathcal{I}}, \ddot{\bm{p}}_d^{\mathcal{I}}$, as the feedforward term along with the desired yaw angle $\psi_d$ for the position controller to track. 
The position controller, in turn, computes the desired attitude $\mathbf{R}_d$ and thrust along the body's z-axis $\bm{T}^{\mathcal{B}}_d$. Based on the desired attitude, the attitude controller calculates the required torque $\bm{M}_d^{\mathcal{B}}$. Ultimately, the Mixer translates $\bm{M}_d^{\mathcal{B}}$ and $\bm{T}^{\mathcal{B}}_d$ into motor speed commands $\Omega_{up}, \Omega_{dw}$ and swashplate servo commands $\delta_{ele,up}$, $\delta_{ail,up}$, $\delta_{ele,dw}$ and $\delta_{ail,dw}$.

\subsubsection{Motion Planning}
In quadrotor trajectory optimization, the principle of differential flatness~\cite{mellinger2011minimum} is widely utilized. Although COMET is equipped with six actuators, the small vertical spacing between its upper and lower rotors, combined with the need to prevent blade interference, restricts its ability to achieve full six-degree-of-freedom control.
As a result, similar to conventional multirotors, COMET is treated as an underactuated vehicle.
This work focuses on near-hover and low-speed flight conditions, where aerodynamic drag is negligible and external forces primarily act along the body’s $z$-axis.
Under these assumptions, the net body force $\bm{F}^{\mathcal{B}}$ can be reasonably approximated as being parallel to $\bm{z}^{\mathcal{B}}$.
This simplification makes COMET’s dynamic behavior analogous to that of traditional quadrotors, allowing the direct application of differential flatness.
Consequently, the existing quadrotor trajectory planning method can be readily adapted for COMET.

\begin{figure*}[t]
    \centering
    \includegraphics[width=1.0\textwidth]{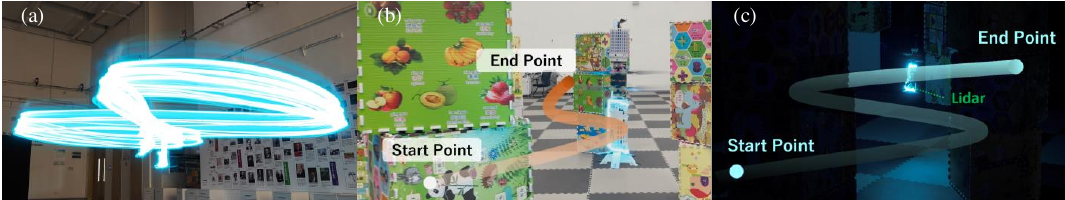}
    \caption{COMET applications in various scenarios. (a).~The composite snapshot demonstrates COMET executing a figure-eight trajectory tracking maneuver. (b).~COMET is autonomously navigating in a well-lit, cluttered environment. (c).~COMET is autonomously navigating in a dimly-lit, cluttered environment.}
    \label{fig:application}
\end{figure*}

\subsubsection{Cascaded Controller}
The COMET's position controller employs a cascaded structure with velocity and acceleration feedforward.
Feedforward terms are derived from the motion planning module, utilizing the first and second derivatives of the reference trajectory as velocity and acceleration inputs, respectively.
The position controller employs a cascaded structure with a proportional gain, calculating the desired velocity. 
The velocity controller computes the desired acceleration $\bm{a}_d^{\mathcal{I}}$ via a PID controller.
The corresponding desired attitude $\mathbf{R}_d$ and thrust force $\bm{T}^{\mathcal{B}}_d$ are determined from the desired acceleration $\bm{a}_d^{\mathcal{I}}$ and the gravity force $m\bm{g}$, following the differential flatness formulation described in~\cite{mellinger2011minimum}: 
\begin{equation}
    m \bm{a}_d^{\mathcal{I}} = m\bm{g} + \mathbf{R}_d \bm{T}^{\mathcal{B}}_d.
\end{equation}
The attitude controller adopts a dual-loop PID control strategy, with an outer loop for attitude control based on quaternion error and an inner loop for angular rate control.

\section{Experiments}
\subsection{Hover Flight Endurance}
Hovering flight time is one of the core performance metrics of the COMET.
To validate the effectiveness of COMET's optimization for efficiency, we conduct endurance tests under hovering conditions with varying take-off weight configurations. 
The experimental results are summarized in Table~\ref{endurance}.
With a battery capacity of 66.6 Wh and a take-off weight of 1250 g, COMET achieves a hover time of 24.3 min and a power efficiency of 7.66 g/W. 
By using a larger battery as an additional payload, COMET can further extend its endurance to 35.5 minutes. 
However, due to the 16.8\% increase in additional load compared to the standard configuration, COMET's efficiency decreases to 7.15 g/W.
By further increasing the additional payload, the take-off weight of COMET is raised to 1940 g, resulting in a 55.2\% increase in additional load compared to the standard configuration, as a result, COMET's efficiency further decreases to 6.00 g/W.
Compared to existing coaxial bi-copters with reported flight efficiency, such as HALO~\cite{li2023halo} and GimbalHawk~\cite{reddington2021design}, as well as commercial multicopter AAVs with the same projected area (DJI Air 2S), COMET demonstrates an advantage in hover efficiency.

The above results demonstrate the effectiveness of COMET's propulsion system design and efficiency optimization. 
COMET is capable of carrying a sufficiently large battery for tasks requiring extended endurance, maintaining a long flight time and high efficiency, while also ensuring that its flight efficiency does not significantly decrease when carrying additional payloads.

\begin{figure*}[t]
    \centering
    \includegraphics[width=1.0\textwidth]{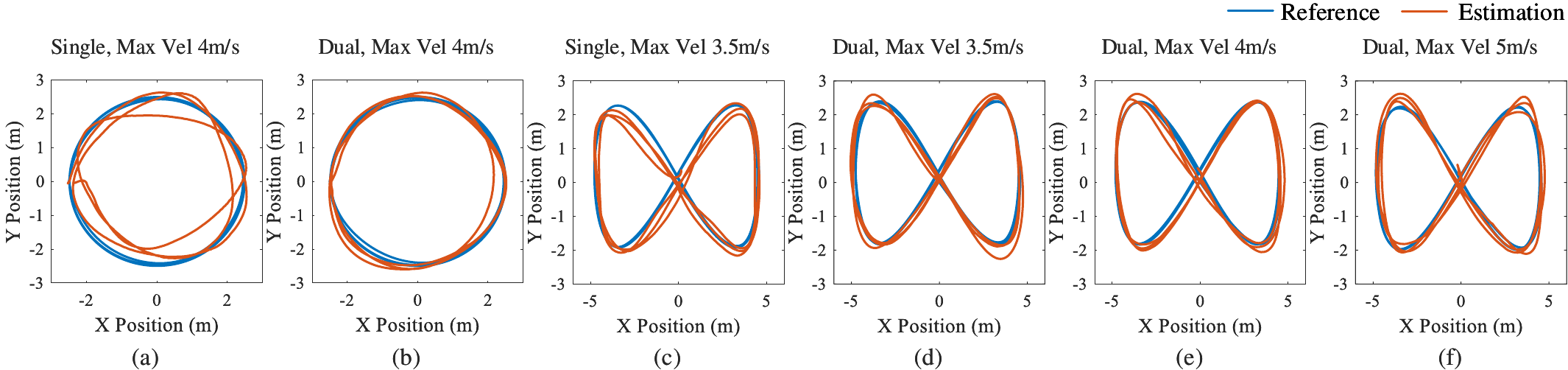}
    \caption{COMET trajectory tracking performance under different swashplate configurations, trajectory types, and maximum velocity. (a)-(b).~COMET tracks a circular trajectory. (c)-(f).~COMET tracks a figure-eight trajectory.}
    \label{fig:traj_result}
    \vspace{-10pt}
\end{figure*}

\begin{table}[h]
  \centering
  \caption{Comparisons of Hover Endurance}
  \label{endurance}
  \renewcommand{\arraystretch}{1.5}
  \begin{tabularx}{\columnwidth}{>{\centering\arraybackslash}p{2cm} >{\centering\arraybackslash}X >{\centering\arraybackslash}X >{\centering\arraybackslash}X >{\centering\arraybackslash}X >{\centering\arraybackslash}X}
    \specialrule{0.2em}{0em}{0.2em}
    \textbf{Prototype} & \textbf{Battery\newline energy\newline(Wh)} & \textbf{Takeoff\newline weight\newline(g)} & \textbf{Hover\newline power\newline(W)} & \textbf{Hover\newline time\newline(min)} & \textbf{Actually\newline Efficient\newline(g/W)} \\
    \midrule
    \textbf{COMET} & \textbf{66.6} & \textbf{1250} & \textbf{163.2} & \textbf{24.3} & \textbf{7.66} \\
    \textbf{COMET} & \textbf{111} & \textbf{1460} & \textbf{204.0} & \textbf{35.5} & \textbf{7.15} \\
    \textbf{COMET} & \textbf{111} & \textbf{1940} & \textbf{294.0} & \textbf{22.0} & \textbf{6.00} \\
    HALO~\cite{li2023halo} & 16.5 & 1344 & 199.5 & N/A & 6.74 \\
    GimbalHawk~\cite{reddington2021design} & 285 & 5380 & 994.3 & N/A & 5.41 \\
    DJI Air 2S & 41.4 & 595 & 101.1 & 24.0 & 5.89 \\
    \specialrule{0.2em}{0.2em}{0em}
  \end{tabularx}
  \vspace{-20pt}
\end{table}

\begin{table*}[t]
	\centering
	\caption{Trajectory Tracking Result of The COMET in Single Swashplate and Dual Swashplate Configurations.}
	\label{tab:comet_track}
	\resizebox{\textwidth}{!}{
		\begin{tabular}{@{}ccccccccc@{}}
			\toprule
			& Single & Dual & Single & Dual & Single & Dual & Dual & Dual\\
            & Circle & Circle & Circle & Circle & Figure eight & Figure eight & Figure eight & Figure eight\\
            \midrule
            Max Vel (m/s) & 3.00 & 3.00 & 4.00 & 4.00 & 3.50 & 3.50 & 4.00 & 5.00 \\
            Max Acc ($m/s^{2}$) & 3.71 & 3.71 & 6.64 & 6.64 & 5.0 & 5.0 & 6.45 & 6.79 \\
			RMSE (m) & 0.2908 & \textbf{0.1712} & 0.3924 & \textbf{0.3139} & 0.3221 & \textbf{0.2534} & 0.2876 & 0.3052\\
            MAE (m) & 0.6161&\textbf{0.3417}&0.8696&\textbf{0.6201}&0.7833 & \textbf{0.5165}&0.5650&0.6526\\
            Min/MAX Roll Control (-) & -0.8013/0.5770&\textbf{-0.3363/0.5214}&-1/0.9893&\textbf{-0.6473/0.5900}&-1/0.8318&\textbf{-0.7554/0.5490}&-0.8821/0.3635&-1/0.4353\\
            Min/Max Pitch Control (-) & -0.6767/0.9912&\textbf{-0.3984/0.6028}&-0.8585/1&\textbf{-0.5806/0.6938}&-0.9308/0.4953&\textbf{-0.6379/0.4239}&-0.4064/0.7926&-0.5734/1\\
            Avg Power (W) & 187.98&\textbf{170.57}&201.86&\textbf{198.39}&194.78&\textbf{186.93}&193.13&197.08\\
            Max Power (W) & 223.22&\textbf{183.75}&262.49&\textbf{257.99}&246.48&\textbf{215.85}&236.77&237.03\\
            Efficiency (g/W) & 7.1283&\textbf{7.8561}&6.6382&\textbf{6.7544}&6.8796&\textbf{7.1686}&6.9382&6.7992\\
			\bottomrule
		\end{tabular}
    }
    \vspace{-10pt}
\end{table*}

\subsection{Trajectory Tracking and Flight Effectiveness}
Trajectory tracking performance is critical for dynamic flight tasks. 
In this experiment, COMET is tasked with tracking a predefined circle trajectory with a diameter of $5\,\mathrm{m}$ and a figure-eight trajectory within a $10\,\mathrm{m} \times 5\,\mathrm{m}$ area, using the NOKOV motion capture system for precise state estimation.
The maximum speed for the generated circular trajectory is set to 3 m/s and 4 m/s, while the maximum speed for the figure-eight trajectory is set to 3.5 m/s, 4 m/s, and 5 m/s.
To evaluate the necessity of the dual swashplate configuration for maneuverability, we include a comparison with a single swashplate version of COMET. 
In this configuration, the cyclic pitch mechanism of the upper rotor is disabled, while all controller parameters are kept identical across both configurations.
Meanwhile, to record the efficiency during dynamic flight missions, a CUAV HV PM power meter is installed on COMET to measure high-precision voltage and current data during flight.
The total take-off weight of COMET in trajectory tracking tasks is 1340g.
Fig.~\ref{fig:application}(a) presents a composite snapshot from one of the trajectory tracking experiments, and the various trajectory tracking results in X and Y positions are shown in Fig.~\ref{fig:traj_result}.
The additional control torque provided by the dual swashplate mechanism enhanced COMET's tracking performance in these tasks, as demonstrated by the comparison between Fig.~\ref{fig:traj_result}(a) and (b), as well as between (c) and (d).
However, in the figure-eight trajectory tracking task with a maximum speed of 4 m/s, COMET with the single swashplate configuration lost control at the first turn due to actuator saturation. 
In contrast, COMET with the dual swashplate configuration is able to effectively track figure-eight trajectories with maximum speeds of 4 m/s and 5 m/s, as shown in Fig.~\ref{fig:traj_result}(e) and (f).
The corresponding quantitative results are summarized in Table~\ref{tab:comet_track}.
In the circular trajectory tracking task with a maximum speed of 3 m/s, COMET with the dual swashplate configuration reduces the root mean square error (RMSE) by 41.12\% and the maximum absolute error (MAE) by 44.54\% compared to the single swashplate configuration. 
In the circular trajectory with a maximum speed of 4 m/s, the RMSE is reduced by 20.01\% and the MAE by 28.69\%. 
In the figure-eight trajectory with a maximum speed of 3.5 m/s, the RMSE is reduced by 21.33\% and the MAE by 34.06\%.
In the figure-eight trajectory tracking task with a maximum speed of 4 m/s, COMET with the single swashplate configuration is unable to complete the flight task. 
However, COMET with the dual swashplate configuration achieves an RMSE of 0.2876 m and an MAE of 0.5650 m. 
Furthermore, COMET with the dual swashplate configuration successfully tracks the figure-eight trajectory with a maximum speed of 5 m/s, achieving an RMSE of 0.3052 m and an MAE of 0.6526 m.
We evaluate the normalized control inputs of COMET's cyclic pitch during flight under two different configurations.
The rate controller outputs are normalized to cyclic pitch control inputs ranging from -1 to 1 through the mixer.
We count the maximum amount of pitch and roll normalized control for different cyclic pitch configurations in the trajectory tracking task. 
As shown in Table.~\ref{tab:comet_track}, the COMET with the single swashplate configuration experiences saturation in the cyclic pitch control inputs for both roll and pitch control (i.e., absolute normalized control inputs reaching 1) in the circular trajectory tracking task with a maximum speed of 4 m/s. 
In contrast, the dual swashplate configuration does not exhibit control input saturation under the maximum flight speed of 5m/s.
These results indicate that the dual swashplate configuration can provide sufficient control torque for agile flight.

Finally, we calculate the average power, maximum power, and efficiency of COMET in the trajectory tracking tasks, as shown in Table~\ref{tab:comet_track}. 
The experimental results indicate that COMET with the dual swashplate configuration, compared to the single swashplate configuration, can reduce power consumption and improve efficiency to some extent during dynamic flight.
In the circular trajectory with a maximum speed of 3 m/s, the dual swashplate configuration reduces the average power by 9.26\% and the maximum power by 17.68\% compared to the single swashplate configuration. 
In the circular trajectory with a maximum speed of 4 m/s, the average power is reduced by 1.72\% and the maximum power by 1.71\%. 
In the figure-eight trajectory with a maximum speed of 3.5 m/s, the average power decreases by 4.03\% and the maximum power by 12.43\%.
This improvement is due to the reduced amplitude of cyclic pitch variation, which decreases the fluctuations in the TPP inclination angle, thereby reducing the fluctuations in the aircraft's thrust. 
As a result, both the average power and maximum power decrease, leading to an improvement in COMET's flight efficiency.

\subsection{Autonomous Navigation with Different Sensing Configurations}
We present several experiments in cluttered, unknown environments to validate the overall system integration of COMET.
We use the Ego-planner~\cite{zhou2020ego} framework for trajectory planning.
The waypoint is provided in advance and is located 15 meters away from the take-off position.
Depending on the lighting conditions, we configure different perception sensors to adapt to the environment.
In well-lit conditions, we use VINS-Fusion~\cite{qin2019general} for visual-inertial state estimation.
In dimly-lit conditions, we use a lidar-inertial odometry (LIO) system, leveraging FAST-LIO2~\cite{xu2022fast} for robust localization.
The snapshots from the flight tests are shown Fig.~\ref{fig:application}(b) and (c).
We refer readers to the supplementary video for more information.
COMET successfully completes autonomous navigation tasks in both conditions, demonstrating the platform’s versatility and robustness across various sensing modalities and environments.

\section{Conclusion and Future Work}
\label{sec:conclusion}
The COMET proposed in this letter features a dual swashplate coaxial bi-copter system.
The vertical separation distance and the blade installation angle of the coaxial bi-copter system are optimized through bench tests, ensuring the optimal rotor system efficiency while maintaining compactness.
Comparative endurance tests confirm that COMET achieves a hover efficiency of up to 7.66 g/W and demonstrates its robustness under varying payload conditions.
Bench tests and comprehensive trajectory tracking experiments, compared to the single swashplate configuration, demonstrate the necessity of the dual swashplate configuration for dynamic flight performance.
Autonomous flight trials further confirm COMET's potential for real-world missions.

However, the current trajectory tracking performance of COMET is constrained by rotor drag and the induced torque disturbances relative to the aircraft's CoG, limiting further improvements in its tracking capability.
Future work will focus on developing a specific trajectory generation and tracking control framework for high-speed flights.

\bibliography{Reference} 

\begin{thebibliography}{10}
\providecommand{\url}[1]{#1}
\csname url@rmstyle\endcsname
\providecommand{\newblock}{\relax}
\providecommand{\bibinfo}[2]{#2}
\providecommand\BIBentrySTDinterwordspacing{\spaceskip=0pt\relax}
\providecommand\BIBentryALTinterwordstretchfactor{4}
\providecommand\BIBentryALTinterwordspacing{\spaceskip=\fontdimen2\font plus
\BIBentryALTinterwordstretchfactor\fontdimen3\font minus \fontdimen4\font\relax}
\providecommand\BIBforeignlanguage[2]{{%
\expandafter\ifx\csname l@#1\endcsname\relax
\typeout{** WARNING: IEEEtran.bst: No hyphenation pattern has been}%
\typeout{** loaded for the language `#1'. Using the pattern for}%
\typeout{** the default language instead.}%
\else
\language=\csname l@#1\endcsname
\fi
#2}}

\bibitem{environmental_monitoring}
V.~Sprincean, A.~Paladi, V.~Andruh, A.~Danici, P.~Lozovanu, and F.~Paladi, ``Uav-based measuring station for monitoring and computational modeling of environmental factors,'' 2021, pp. 80--85.

\bibitem{disaster}
L.~Qian, Y.~L. Lo, and H.~H.~T. Liu, ``Energy aware planning of a quadrotor with 5g access point in disaster response,'' in \emph{2024 IEEE 10th World Forum on Internet of Things (WF-IoT)}, 2024.

\bibitem{exploration}
B.~Zhou, H.~Xu, and S.~Shen, ``Racer: Rapid collaborative exploration with a decentralized multi-uav system,'' \emph{IEEE Transactions on Robotics}, 2023.

\bibitem{infrastructur_inspection}
M.~Imad, M.~Wicaksono, and S.~Y. Shin, ``Autonomous uav navigation and real-time crack detection for infrastructure inspection in gps-denied environments,'' 2024, pp. 1210--1213.

\bibitem{uav_2018}
T.~Rakha and A.~Gorodetsky, ``Review of unmanned aerial system (uas) applications in the built environment: Towards automated building inspection procedures using drones,'' \emph{Automation in Construction}, 2018.

\bibitem{Koe2012single}
A.~Koehl, H.~Rafaralahy, M.~Boutayeb, and B.~Martinez, ``\BIBforeignlanguage{en-US}{Aerodynamic modelling and experimental identification of a coaxial-rotor uav},'' \emph{\BIBforeignlanguage{en-US}{Journal of Intelligent and Robotic Systems}}, p. 53–68, Sep 2012.

\bibitem{Koehl_Rafaralahy_Martinez_Boutayeb_2010}
A.~Koehl, H.~Rafaralahy, B.~Martinez, and M.~Boutayeb, ``Modeling and identification of a launched micro air vehicle: Design and experimental results,'' \emph{AIAA Modeling and Simulation Technologies Conference and Exhibit,AIAA Modeling and Simulation Technologies Conference and Exhibit}, 2010.

\bibitem{buzzatto2022benchmarking}
J.~Buzzatto and M.~Liarokapis, ``A benchmarking platform and a control allocation method for improving the efficiency of coaxial rotor systems,'' \emph{IEEE Robotics and Automation Letters}, vol.~7, no.~2, pp. 5302--5309, 2022.

\bibitem{bermes2008CoG}
C.~Bermes, S.~Leutenegger, S.~Bouabdallah, D.~Schafroth, and R.~Siegwart, ``\BIBforeignlanguage{en-US}{New design of the steering mechanism for a mini coaxial helicopter},'' in \emph{\BIBforeignlanguage{en-US}{2008 IEEE/RSJ International Conference on Intelligent Robots and Systems}}, Sep 2008.

\bibitem{canfly}
N.~Pan, R.~Jin, C.~Xu, and F.~Gao, ``Canfly: A can-sized autonomous mini coaxial helicopter,'' in \emph{2023 IEEE/RSJ International Conference on Intelligent Robots and Systems (IROS)}, 2023, pp. 4989--4996.

\bibitem{li2023halo}
H.~Li, N.~Chen, F.~Kong, Y.~Zou, S.~Zhou, D.~He, and F.~Zhang, ``Halo: A safe, coaxial, and dual-ducted uav without servo,'' in \emph{2023 IEEE/RSJ International Conference on Intelligent Robots and Systems (IROS)}.\hskip 1em plus 0.5em minus 0.4em\relax IEEE, 2023, pp. 6935--6941.

\bibitem{chen2024design}
L.~Chen, J.~Xiao, Y.~Zheng, N.~A. Alagappan, and M.~Feroskhan, ``Design, modeling, and control of a coaxial drone,'' \emph{IEEE Transactions on Robotics}, vol.~40, pp. 1650--1663, 2024.

\bibitem{reddington2021design}
J.~Reddington, ``Design, development, and tuning of a gimbaled coaxial uav,'' 2021.

\bibitem{fank2011}
P.~Fankhauser, S.~Bouabdallah, S.~Leutenegger, and R.~Siegwart, ``\BIBforeignlanguage{en-US}{Modeling and decoupling control of the coax micro helicopter},'' in \emph{\BIBforeignlanguage{en-US}{2011 IEEE/RSJ International Conference on Intelligent Robots and Systems}}, Sep 2011.

\bibitem{boua2006CoG}
S.~Bouabdallah, R.~Siegwart, and G.~Caprari, ``Design and control of an indoor coaxial helicopter,'' in \emph{IEEE/RSJ International Conference on Intelligent Robots and Systems}, 2006.

\bibitem{wang2006CoG}
W.~Wang, G.~Song, K.~Nonami, M.~Hirata, and O.~Miyazawa, ``Autonomous control for micro-flying robot and small wireless helicopter xrb,'' in \emph{2006 IEEE/RSJ International Conference on Intelligent Robots and Systems}.\hskip 1em plus 0.5em minus 0.4em\relax IEEE, 2006, pp. 2906--2911.

\bibitem{deng2020aerodynamic}
S.~Deng, S.~Wang, and Z.~Zhang, ``Aerodynamic performance assessment of a ducted fan uav for vtol applications,'' \emph{Aerospace Science and Technology}, vol. 103, p. 105895, 2020.

\bibitem{dominguez2022micro}
V.~H. Dominguez, O.~Garcia-Salazar, L.~Amezquita-Brooks, L.~A. Reyes-Osorio, C.~Santana-Delgado, and E.~G. Rojo-Rodriguez, ``Micro coaxial drone: flight dynamics, simulation and ground testing,'' \emph{Aerospace}, vol.~9, no.~5, p. 245, 2022.

\bibitem{Paulos2018}
J.~Paulos, B.~Caraher, and M.~Yim, ``\BIBforeignlanguage{en-US}{Emulating a fully actuated aerial vehicle using two actuators},'' in \emph{\BIBforeignlanguage{en-US}{2018 IEEE International Conference on Robotics and Automation (ICRA)}}, May 2018.

\bibitem{gun_2013}
E.~Roussel, P.~Gnemmi, and S.~Changey, ``\BIBforeignlanguage{en-US}{Gun-launched micro air vehicle: Concept, challenges and results},'' in \emph{\BIBforeignlanguage{en-US}{2013 International Conference on Unmanned Aircraft Systems (ICUAS)}}, May 2013.

\bibitem{qin_2024}
Z.~Qin, J.~Wei, M.~Cao, B.~Chen, K.~Li, and K.~Liu, ``Design and flight control of a novel tilt-rotor octocopter using passive hinges,'' \emph{IEEE Robotics and Automation Letters}, 2024.

\bibitem{zhao_2022}
C.~Zhao, H.~Lin, L.~Feng, S.~Bai, F.~Hu, and L.~Zhao, ``Backstepping-based tracking control for a coaxial rotor uav,'' in \emph{2022 China Automation Congress (CAC)}, 2022.

\bibitem{KimBrown2008}
H.~D. Kim and R.~E. Brown, ``Modelling the aerodynamics of coaxial helicopters,'' in \emph{26th AIAA Applied Aerodynamics Conference}, 2008.

\bibitem{SwashplateControlUAV}
V.~Hugar, P.~Matt, C.~Chandrashekar, and K.~M, ``Modeling and simulation of helicopter swashplate collective control response in uav applications,'' in \emph{2024 8th International Conference on Computational System and Information Technology for Sustainable Solutions (CSITSS)}, 2024.

\bibitem{Rama2013aero}
M.~Ramasamy, ``\BIBforeignlanguage{en-US}{Measurements comparing hover performance of single, coaxial, tandem, and tilt-rotor configurations},'' \emph{\BIBforeignlanguage{en-US}{AHS 69th Annual Forum}}, May 2013.

\bibitem{Hai2018aero}
W.~Hai, H.~Qiang, Z.~Yanchang, Y.~Chunlai, and W.~Li, ``\BIBforeignlanguage{en-US}{Research on the influence of coaxial rotor configuration on the propulsion of multi-rotor uavs},'' in \emph{\BIBforeignlanguage{en-US}{2018 IEEE 13th Annual International Conference on Nano/Micro Engineered and Molecular Systems (NEMS)}}, Apr 2018.

\bibitem{lak2010areo}
V.~K. Lakshminarayan and J.~D. Baeder, ``\BIBforeignlanguage{en-US}{Computational investigation of microscale shrouded rotor aerodynamics in hover},'' \emph{\BIBforeignlanguage{en-US}{Journal of the American Helicopter Society}}, p. 1–15, Oct 2011.

\bibitem{ramasamy2015hover}
M.~Ramasamy, ``Hover performance measurements toward understanding aerodynamic interference in coaxial, tandem, and tilt rotors,'' \emph{Journal of the American Helicopter Society}, vol.~60, no.~3, pp. 1--17, 2015.

\bibitem{wang_2012_dynamic}
F.~Wang, S.~K. Phang, J.~Cui, G.~Cai, B.~M. Chen, and T.~H. Lee, ``Nonlinear modeling of a miniature fixed-pitch coaxial uav,'' in \emph{2012 American Control Conference (ACC)}, 2012.

\bibitem{Giljarhus_2022}
K.~E.~T. Giljarhus, A.~Porcarelli, and J.~Apeland, ``\BIBforeignlanguage{en-US}{Investigation of rotor efficiency with varying rotor pitch angle for a coaxial drone},'' \emph{\BIBforeignlanguage{en-US}{Drones}}, p.~91, Apr 2022.

\bibitem{mellinger2011minimum}
D.~Mellinger and V.~Kumar, ``Minimum snap trajectory generation and control for quadrotors,'' in \emph{2011 IEEE international conference on robotics and automation}.\hskip 1em plus 0.5em minus 0.4em\relax IEEE, 2011, pp. 2520--2525.

\bibitem{zhou2020ego}
X.~Zhou, Z.~Wang, H.~Ye, C.~Xu, and F.~Gao, ``Ego-planner: An esdf-free gradient-based local planner for quadrotors,'' \emph{IEEE Robotics and Automation Letters}, vol.~6, no.~2, pp. 478--485, 2020.

\bibitem{qin2019general}
T.~Qin, J.~Pan, S.~Cao, and S.~Shen, ``A general optimization-based framework for local odometry estimation with multiple sensors,'' \emph{arXiv preprint arXiv:1901.03638}, 2019.

\bibitem{xu2022fast}
W.~Xu, Y.~Cai, D.~He, J.~Lin, and F.~Zhang, ``Fast-lio2: Fast direct lidar-inertial odometry,'' \emph{IEEE Transactions on Robotics}, vol.~38, no.~4, pp. 2053--2073, 2022.

\end{thebibliography}
\end{document}